\documentclass{article}

\usepackage{times}
\usepackage{graphicx} 

\usepackage{natbib}

\usepackage{algorithm}
\usepackage{algorithmic}
\usepackage{epsfig}
\usepackage{amsmath}
\usepackage{amsfonts}
\usepackage{epstopdf}

\usepackage[accepted]{icml2015}

\icmltitlerunning{}

\begin{document} 

\twocolumn[
\icmltitle{Long Short-Term Memory Over Tree Structures}

\icmlauthor{Xiaodan Zhu}{Xiaodan.Zhu@nrc-cnrc.gc.ca}
\icmladdress{National Research Council Canada,
            1200 Montreal Road M-50, Ottawa, ON K1A 0R6 CANADA}
\icmlauthor{Parinaz Sobhani}{psobh090@uottawa.ca}
\icmladdress{School of Electrical Engineering and Computer Science, University of Ottawa,
            800 King Edward Avenue, Ottawa, ON K1N 6N5 CANADA}
\icmlauthor{Hongyu Guo}{Hongyu.Guo@nrc-cnrc.gc.ca}
\icmladdress{National Research Council Canada,
            1200 Montreal Road M-50, Ottawa, ON K1A 0R6 CANADA}

\icmlkeywords{boring formatting information, machine learning, ICML}

\vskip 15mm
]

\begin{abstract} 
The chain-structured long short-term memory (LSTM) has showed to be effective in a wide range of problems such as speech recognition and machine translation. In this paper, we propose to extend it to tree structures, in which a memory cell can reflect the history memories of multiple \textit{child cells} or multiple \textit{descendant cells} in a recursive process. We call the model S-LSTM, which provides a principled way of considering long-distance interaction over hierarchies, e.g., language or image parse structures. We leverage the models for semantic composition to understand the meaning of text, a fundamental problem in natural language understanding, and show that it outperforms a state-of-the-art recursive model by replacing its composition layers with the S-LSTM memory blocks. We also show that utilizing the given structures is helpful in achieving a performance better than that without considering the structures. 
\end{abstract} 


\section{Introduction}
\label{sec:introduction}
Recent years have seen a revival of the long short-term memory (LSTM)~\cite{Hochreiter1997}, with its effectiveness being demonstrated on a wide range of  problems such as speech recognition \cite{Graves:2013},  machine translation~\cite{Sutskever:2014,Cho:2014}, and image-to-text conversion \cite{Vinyals:2014}, among many others, in which history is summarized and coded in the \textit{memory cell} in a \textit{full-order} time sequence.


Recursion is a fundamental process associated with many problems---a recursive process and hierarchical structure so formed are common in different modalities. For example, semantics of sentences in human languages is believed to be carried by not merely a linear concatenation of words; instead, sentences have parse structures \cite{Manning:1999}. Image understanding, as another example, benefits from recursive modeling over structures, which yielded the state-of-the-art performance on tasks like scene segmentation~\cite{Socher2011b}. 

In this paper, we extend LSTM to tree structures, in which we learn memory cells that can reflect the history memories of multiple \textit{child cells} and hence multiple \textit{descendant cells}. We call the model S-LSTM. 
Compared with previous recursive neural networks \cite{Socher2013,Socher2012}, S-LSTM has the potentials of avoiding \textit{gradient vanishing} and hence may model long-distance interaction over trees. This is a desirable characteristic as many of such structures are deep. S-LSTM can be considered as bringing the merits of a recursive neural network and a recurrent neural network together\footnote{As both of them can be shortened to be RNN, in the rest of this paper we refer to a Recurrent Neural Network as RNN and a Recursive Neural Network as RvNN.}. In short, S-LSTM wires memory blocks in a partial-order structures instead of in a full-order sequence as in a chain-structured LSTM.

We leverage the S-LSTM model to solve a semantic composition problem that learns the meaning for a piece of texts---learning good representations for meaning of text is core to automatically understanding human languages. More specifically, we experiment with the models on the Stanford Sentiment Tree Bank~\cite{Socher2013} to determine the sentiment for different granularities of phrases in a tree. The dataset has favorable properties: in addition to being a benchmark for much previous work, it provides with human annotations at all nodes of the trees, enabling us to comprehensively explore the properties of S-LSTM. We experimentally show that S-LSTM outperforms a state-of-the-art recursive model by simply replacing the original tensor-enhanced composition with the S-LSTM memory block we propose here. We showed that utilizing the given structures is helpful in achieving a better performance than that without considering the structures. 

\section{Related Work}
{\bfseries Recursive neural networks}
Recursion is a fundamental process in different modalities. In recent years, recursive neural networks (RvNN) have been introduced and demonstrated to achieve state-of-the-art performances on different problems such as semantic analysis in natural language processing and image segmentation~\cite{Socher2013,Socher2011b}. These networks are defined over recursive tree structures---a tree node is a vector computed from its children. In a recursive fashion, the information from the leaf nodes of a tree and its internal nodes are combined in a bottom-up manner through the tree. Derivatives of errors are computed with backpropagation over structures~\cite{Goller96learningtask-dependent}.


In addition, the literature has also included many other efforts of applying feedforward-based neural network over structures, including~\cite{Goller96learningtask-dependent,CHATER92,Starzyk_anticipation-basedtemporal,Hammer04ageneral}, amongst others. For instance, Legrand and Collobert leverage neural networks over greedy syntactic parsing~\cite{pinheiro:2014}.  In~\cite{Irsoy2014}, a deep recursive neural network is proposed .
Nevertheless, over the often deep structures, the networks are potentially subject to the vanishing gradient problem, resulting in difficulties in leveraging long-distance dependencies in the structures. In this paper, we propose the S-LSTM model that wires memory blocks in recursive structures. We compare our model with the RvNN models presented in ~\cite{Socher2013}, as we directly replaced the tensor-enhanced composition layer at each tree node with a S-LSTM memory block. We show the advantages of our proposed model in achieving significantly better results. 

{\bfseries Recurrent neural networks and LSTM}
Unlike a feed-forward network, a recurrent neural network (RNN) shares their hidden states across time. The sequential history is summarized in a hidden vector. RNN also suffers from the decaying of gradient, or less frequently, blowing-up of gradient problem. LSTM replaces the hidden vector of a recurrent neural network with \textit{memory blocks} which are equipped with gates; it can in principle keep long-term memory by training proper gating weights (refer to \cite{Graves2008} for intuitive illustrations and good discussions), and it has practically showed to be very useful, achieving the state of the art on a range of problems including speech recognition~\cite{Graves:2013}, digit handwriting recognition~\cite{Liwicki07anovel,graves2012supervised}, and achieve interesting results on statistical machine translation~\cite{Sutskever:2014,Cho:2014} and music composition~\cite{Eck02learningthe,Eck02findingtemporal}. In~\cite{Graves:2013}, a deep LSTM network  achieved the state-of-the-art results on the TIMIT phoneme recognition benchmark. In~\cite{Sutskever:2014,Cho:2014}, a pair of LSTM networks are trained to encode and decode human language for automatic machine translation, which is in particular effective for the more challenging long sentence translation.  
In~\cite{Liwicki07anovel,graves2012supervised}, LSTM networks are found to be very useful for digit writing recognition because of the network's capability of memorizing context information in a long sequence. 
In~\cite{Eck02learningthe,Eck02findingtemporal}, LSTM networks are trained to effectively capture global structures of the temporal data. With the memory cells, LSTM is able to keep track of temporally distant events that indicate global music structures. As a result, LSTM can be successfully trained to compose music, where other RNNs have failed to do so. 

Although promising results have been observed by applying chain-structured LSTM, many other interesting problems are inherently associated with input structures that are more complicated than a sequence. For example, sentences in human languages are believed to be carried by not merely a linear sequence of words; instead, meaning is thought to interweave with structures. While a sequential application of LSTM may capture structural information implicitly, in practice it sometimes lacks the claimed power. For example, even simply reversing the input sequences may result in significant differences in modeling performances, in tasks such as machine translation and speech recognition. Unlike in previous work, we propose here to directly wire memory blocks in recursive structures. We show the proposed S-LSTM model does utilize the structures and achieve results better than those ignoring such priori structures.


\section{The Model}
\begin{figure}[ht]
\centerline{\includegraphics[width=6cm, height=7.3cm]{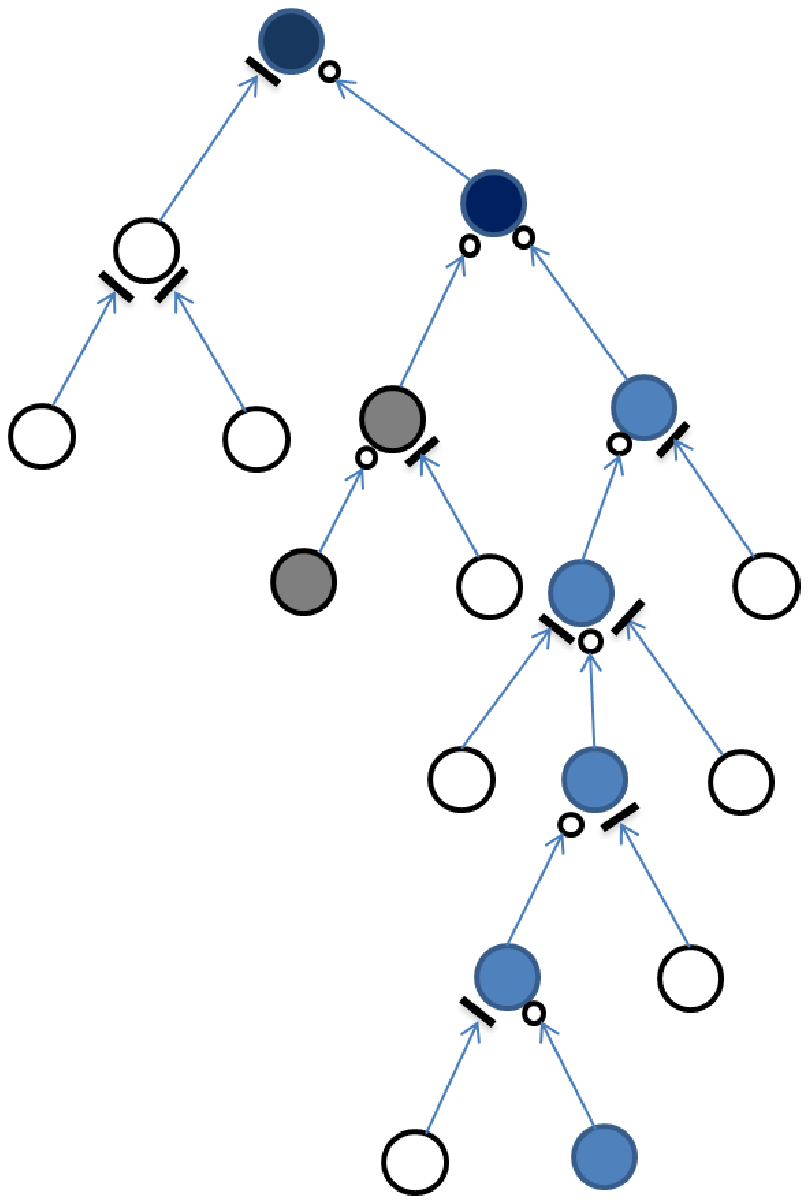}}
\caption{An example of S-LSTM, a long-short term memory network on tree structures. A tree node can consider information from multiple descendants. Information of the other nodes in white are blocked. The small circle ("$\circ$") or short line ("$-$") at each arrowhead indicates a \textit{pass} or \textit{block} of information, respectively, while in the real model the gating is a soft version of gating.}
\label{fig:lstm}
\end{figure}

\noindent \textbf{Model brief} In this paper, we extend LSTM to structures, in which a memory cell can reflect the history memories of multiple child cells and hence multiple descendant cells in a hierarchical structure. As intuitively showed in Figure~\ref{fig:lstm}, the root of the tree can in principle consider information from long-distance interactions over the tree---in this figure, the gray and light-blue leaf. In the figure, the small circle ("$\circ$") or short line ("$-$") at each arrowhead indicates \textit{pass} and \textit{block} of information, respectively. Note that the figure shows a binary case, while in real models a soft version of gating is applied, where a gating signal is in the range of [0, 1], often enforced with a logistic sigmoid function. Through learning the gating signals, as detailed later in this section, S-LSTM provides a principled way of considering long-distance interplays over the input structures. 

\begin{figure}[ht]
\vskip 0.2in
\begin{center}
\centerline{\includegraphics[width=8.5cm, height=8.3cm]{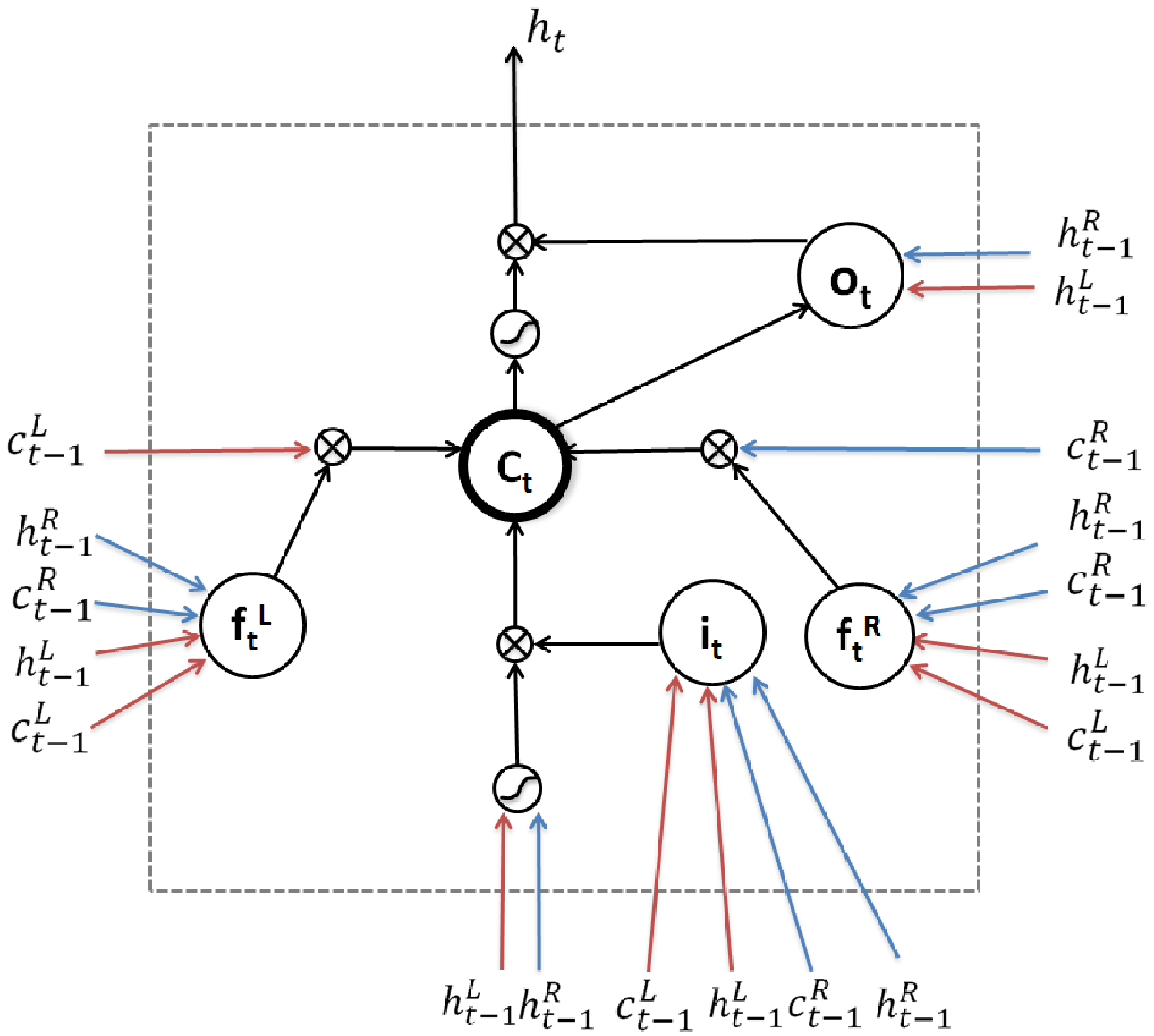}}
\caption{A S-LSTM memory block, consisting of an input gate, two forget gates, and an output gate. Hidden vectors $h_{t-1}^*$ and cell vectors $c_{t-1}^*$ from the left (red arrows) and right (blue arrows) children are deployed to compute $c_t$ and $h_t$. $\otimes$ denotes a Hadamard product, and the ``s" shaped sign is a squashing function (in this paper the \textit{tanh} function). }
\label{fig:model_1}
\end{center}
\end{figure}

\noindent \textbf{The memory block}
Each node in Figure~\ref{fig:lstm} is composed of a S-LSTM \textit{memory block}. We present a specific wiring of such a block in Figure \ref{fig:model_1}. Each memory block contains one input gate and one output gate. The number of forget gates depends on the structure, i.e., the number of children of a node. In this paper, we assume there are two children at each nodes, same as in \cite{Socher2013} and therefore we use their data in our experiments. That is, we have two forget gates. Extension of the model to handle more children is rather straightforward.

As shown in the figure, the hidden vectors of the two children, denoted as $h_{t-1}^L$ for the left child and $h_{t-1}^R$ for the right, are taken in as input of the current block. The input gate $i_t$ consider four resources of information: the hidden vectors ($h_{t-1}^L$ and $h_{t-1}^R$) and cell vectors ($c_{t-1}^L$ and $c_{t-1}^R$) of its two children. These four sources of information are also used to form the gating signals for the left forget gate $f_{t-1}^L$ and right forget gate $f_{t-1}^R$, where the weights used to combining them are specific to each of these gates, denoted as different $W$ in the formulas below. Different from the process in a regular LSTM, the cell here considers the copies from both children's cell vectors ($c_{t-1}^L$, $c_{t-1}^R$), gated with separated forget gates. The left and right forget gates can be controlled independently, allowing the pass-through of information from children's cell vectors. The output gate $o_t$ considers the hidden vectors from the children and the current cell vector. In turn, the hidden vector $h_t$ and the cell vector $c_t$ of the current block are passed to the parent and are used depending on if the current block is a left or right child of its parent. 
In this way, the memory cell, through merging the gated cell vectors of the children, can reflect multiple direct or indirect descendant cells. As a result, the long-distance interplays over the structures can be captured. More specifically, the forward computation of a S-LSTM memory block is specified in the following equations. 
\begin{align}
\label{formula:forward_begin}
\nonumber
 i_t&= \sigma(W_{hi}^L h_{t-1}^L+W_{hi}^R h_{t-1}^R +W_{ci}^L c_{t-1}^L\\ 
 &+ W_{ci}^R c_{t-1}^R+b_i) 
\end{align}
\begin{align}
\nonumber
f_t^L&= \sigma(W_{{hf}_l}^L h_{t-1}^L+W_{{hf}_l}^R h_{t-1}^R+ W_{{cf}_l}^L c_{t-1}^L\\
&+W_{{cf}_l}^R c_{t-1}^R+b_{f_l})\\[10pt]
\nonumber
f_t^R&= \sigma(W_{{hf}_r}^L h_{t-1}^L+W_{{hf}_r}^R h_{t-1}^R+ W_{{cf}_r}^L c_{t-1}^L\\
&+W_{{cf}_r}^R c_{t-1}^R+b_{f_r})\\[10pt]
x_t&= W_{hx}^L h_{t-1}^L +W_{hx}^R h_{t-1}^R+ b_x \\[10pt]
c_t&= f_t^L \otimes c_{t-1}^L+ f_t^R \otimes c_{t-1}^R+ i_t \otimes tanh (x_t)\\[10pt]
\label{formula:outputGate}
o_t&= \sigma(W_{ho}^L h_{t-1}^L+W_{ho}^R h_{t-1}^R+W_{co} c_{t} + b_o)\\[10pt]
 h_t&= o_t \otimes \tanh (c_t)
\label{formula:forward_end}
\end{align}

where $\sigma$ is the element-wise logistic function used to confine the gating
signals to be in the range of [0, 1]; ${f^L}$ and ${f^R}$ are the left and right forget gate, respectively; $b$ is bias and $W$ is network weight matrices; the sign $\otimes$ is a Hadamard product, i.e., element-wise product. The subscripts of the weight matrices indicate what they are used for. For example, $W_{ho}$ is a matrix mapping a hidden vector to an output gate.

\noindent \textbf{Backpropagation over structures}   
During training, the gradient of the objective function with respect to each parameter can be calculated efficiently via backpropagation
over structures~\cite{Goller96learningtask-dependent,Socher2013}. 
The major difference from that of \cite{Socher2013} is we use LSTM-like backpropagation,
where unlike a regular LSTM, pass of error 
needs to discriminate between the left and right children, or in a topology with more than two children, needs to
discriminate between children. Obtaining the 
backprop formulas is tedious but we list them below to facilitate duplication
of our work~\footnote{The code will be published at www.icml-placeholder-only.com}. 
We will discuss the specific objective function later in experiments.
For each memory block, assume that the error passed to the hidden vector is $\epsilon_t^h$.
The derivatives of the output gate $\delta_{t}^o$, left forget gate $\delta_{t}^{f_l}$, 
right forget gate $\delta_{t}^{f_r}$, and input gate $\delta_{t}^i$ are computed as:


\vskip -0.15in
\begin{align}
\epsilon_t^h&= \frac{\partial O}{\partial h_t}\\[3pt]
\delta_{t}^o&= \epsilon_t^h \otimes \tanh(c_t) \otimes \sigma'(o_t)\\[3pt]
\delta_{t}^{f_l}&= \epsilon_t^c \otimes c_{t-1}^L \otimes \sigma'(f_t^L)  \\[3pt]
\delta_{t}^{f_r}&= \epsilon_t^c \otimes c_{t-1}^R \otimes \sigma'(f_t^R)  \\[3pt]
\delta_{t}^i&= \epsilon_t^c \otimes \tanh(x_t) \otimes \sigma'(i_t)   
\end{align}
\vskip 0.1in

where ${\sigma'(x)}$ is the element-wise derivative of the logistic function over vector $x$.
Since it can be computed with the activation of $x$, we abuse the 
notation a bit to write it over the activated vectors in these equations. $\epsilon_t^c$
is the derivative over the cell vector. So if the current node is the left child of its parent, we use Equation (\ref{formula:backleft}) to calculate $\epsilon_t^c$, 
otherwise Formula (\ref{formula:backright}) is used:
\vskip 0.05in
\begin{align}
\label{formula:backleft}
\nonumber
\epsilon_t^c= &\epsilon_t^h \otimes o_t \otimes g'(c_t) + \epsilon^c_{t+1} \otimes f_{t+1}^L +\\[3pt] \nonumber
&(W_{ci}^L)^T\delta_{t+1}^{i} + (W_{cf_l}^L)^T\delta_{t+1}^{f_l}+\\[3pt]
&(W_{cf_r}^L)^T\delta_{t+1}^{f_r} + (W_{co})^T\delta^o_{t} 
\end{align}
\begin{align}
\label{formula:backright}
\nonumber
\epsilon_t^c= &\epsilon_t^h \otimes o_t \otimes g'(c_t) + \epsilon^c_{t+1} \otimes f_{t+1}^R + \\[3pt] \nonumber
&(W_{ci}^R)^T\delta_{t+1}^{i} + (W_{cf_l}^R)^T\delta_{t+1}^{f_l}+\\[3pt]
&(W_{cf_r}^R)^T\delta_{t+1}^{f_r} + (W_{co})^T\delta^o_{t} 
\end{align}
\vskip 0.1in
where ${g'(x)}$ is the element-wise derivative of the \textit{tanh} function. It can also be directly calculated from 
the \textit{tanh} activation of $x$. The superscript $T$ over the weight matrices means matrix transpose.

With derivatives at each gate computed, the derivatives of the weight matrices used in 
Formula (\ref{formula:forward_begin})-(\ref{formula:forward_end}) can be calculated accordingly, 
which is omitted here. We checked the correctness of the S-LSTM implementation with the standard approximated gradient approach.


\noindent \textbf{Objective over trees} 
The objective function defined over structures can be complicated, which could consider the output structures depending on the properties of problem. Following \cite{Socher2013}, the overall objective function we used to learn S-LSTM in this paper is simply minimizing the overall cross-entropy errors and a sum of that at all nodes. 


\section{Experiment Set-up}
As discussed earlier, recursion is a basic process inherent to many problems. 
In this paper, we leverage the proposed model to solve semantic composition for the meanings of pieces of text, a fundamental problem in understanding human languages.


We specifically attempt to determine the sentiment of different granularities of phrases in a tree, within the Stanford Sentiment Tree Bank benchmark data ~\cite{Socher2013}. In obtaining the sentiment of a long piece of text, early work often factorized the problem to consider smaller pieces of component words or phrases with bag-of-words or bag-of-phrases models~\cite{PangL08,Liu12}. More recent work has started to model composition~\cite{moilanen07,Choi08,Socher2012,Socher2013,Kalchbrenner2014}, a more principled approach to modeling the formation of semantics. In this paper, we put the proposed LSTM memory blocks at tree nodes---we replaced the tensor-enhanced composition layer at each tree node presented in ~\cite{Socher2013} with a S-LSTM memory block. We used the same dataset, the Stanford Sentiment Tree Bank, to evaluate the performances of the models. In addition to being a benchmark for much previous work, the data provide with human annotations at all nodes of the trees, facilitating a more comprehensive exploration of the properties of S-LSTM. 


\subsection{Data Set} 

The Stanford Sentiment
Tree Bank~\cite{Socher2013} contains about 
11,800 sentences from the movie reviews that 
were originally discussed in \cite{Pang2005}. The sentences were parsed
with the Stanford parser \cite{Klein2003}. Phrases at all the tree
nodes were manually annotated with sentiment values. We use the same split of
the training and test data as in~\cite{Socher2013} to predict the sentiment categories 
of the roots (sentences) and all phrases (including sentences). For the root sentiment, the training, development, and test sentences are 8544, 1101, and 2210, respectively. The phrase sentiment task includes 318582, 41447, and 82600 phrases for the three sets. Following~\cite{Socher2013}, we also use the 
 classification accuracy to measure the performances.


\subsection{Training Details} 
As mentioned before, we follow \cite{Socher2013} to minimize the cross-entropy error for all nodes or for roots only, depending on specific experiment settings.  
For all phrases, the error is calculated as a regularized sum:

\vspace{1mm}
\begin{align}
\label{err}
E(\theta)= \sum_{i} \sum_{j} t_{j}^{i}\text{log}{y^{{sen_i}}}_{j} + \lambda \left\| \theta \right\|^{2}_{2}
\end{align}
\noindent where $y^{{sen_i}} \in \mathbb{R}^{c \times 1}$ is predicted distribution and $t^{i} \in \mathbb{R}^{c \times 1}$ the target distribution. $c$ is the number of classes or categories, and  $j \in c$ denotes the $j$-th element of the multinomial target distribution; $i$ iterates over nodes, $\theta$ are model parameters, and $\lambda$ is  a regularization parameter. We tuned our model against the development data set as split in~\cite{Socher2013}. 


\label{sec:results} 
\section{Results}

To understand the modeling advantages of 
S-LSTM over the structures, we conducted four sets of experiments. 

\noindent \textbf{Default setting} In the default setting, we conducted experiments as in~\cite{Socher2013}.
Table \ref{tab:main} shows the accuracies of different models on the test set of the Stanford Sentiment Tree Bank. We present the results on 5-category sentiment prediction at both the sentence level (i.e., the \textit{ROOTS} column in the table) and for all phrases including roots (the \textit{PHRASES} column) \footnote{The Stanford CoreNLP package (http://nlp.stanford.edu/sentiment/code.html) only gives approximate accuracies for 2-category sentiment, which are not included here in the table.}. In Table \ref{tab:main}, \textit{NB} and \textit{SVM} are naive Bayes and support vector machine classifiers, respectively; \textit{RvNN} corresponds to RNN in~\cite{Socher2013}. As described earlier, we refer to recursive neural networks to as RvNN to avoid confusion with recurrent neural networks. RNTN is different from RvNN in that when merging two nodes to obtain the hidden vector of their parent, tensor is used to obtain the second-degree
polynomial interactions. 


\begin{table}[h]
\caption{Performances (accuracies) of different models on the test set of Stanford Sentiment Tree Bank, at the sentence level (roots) and the phrase level. $\dag$ shows the performance are statistically significantly better ($p<0.05$) than the corresponding models.} 
\label{tab:main}
\vskip 0.15in
\begin{center}
\begin{small}
\begin{sc}
\begin{tabular}{llllllll}
\noalign{\hrule height 1pt}
\abovespace\belowspace
Models &&&&& roots && phrases\\ 
\hline
\abovespace
\abovespace

NB &&&&& 41.0  && 67.2 \\ 
SVM &&&&& 40.7 && 64.3\\ 
RvNN &&&&& 43.2  && 79.0 \\ 
RNTN &&&&& 45.7 && 80.7\\ 
\belowspace
S-LSTM &&&&& \textbf{48.0}\dag && \textbf{81.9}\dag\\ 
\noalign{\hrule height 1pt}
\end{tabular}
\end{sc}
\end{small}
\end{center}
\vskip 0.1in
\end{table}

Table \ref{tab:main} showed that S-LSTM achieved the best predictive performance, when compared to all the models reported
in~\cite{Socher2013}. The S-LSTM results reported here were obtained by setting the size of the hidden units to be 100, batch size to be 10, and learning rate to be 0.1. In our experiments, we only tuned these hyper-parameters, and we feel that more finer tuning, such as discriminating the classification weights between the leaves (word embedding) and other nodes, using different numbers of hidden units for the memory blocks (e.g., for the hidden layers of words), or different initializations of word embedding, may further improve the performances reported here. 

\begin{figure}[ht]
\vskip 0.2in
\centerline{\includegraphics[width=7cm, height=8cm]{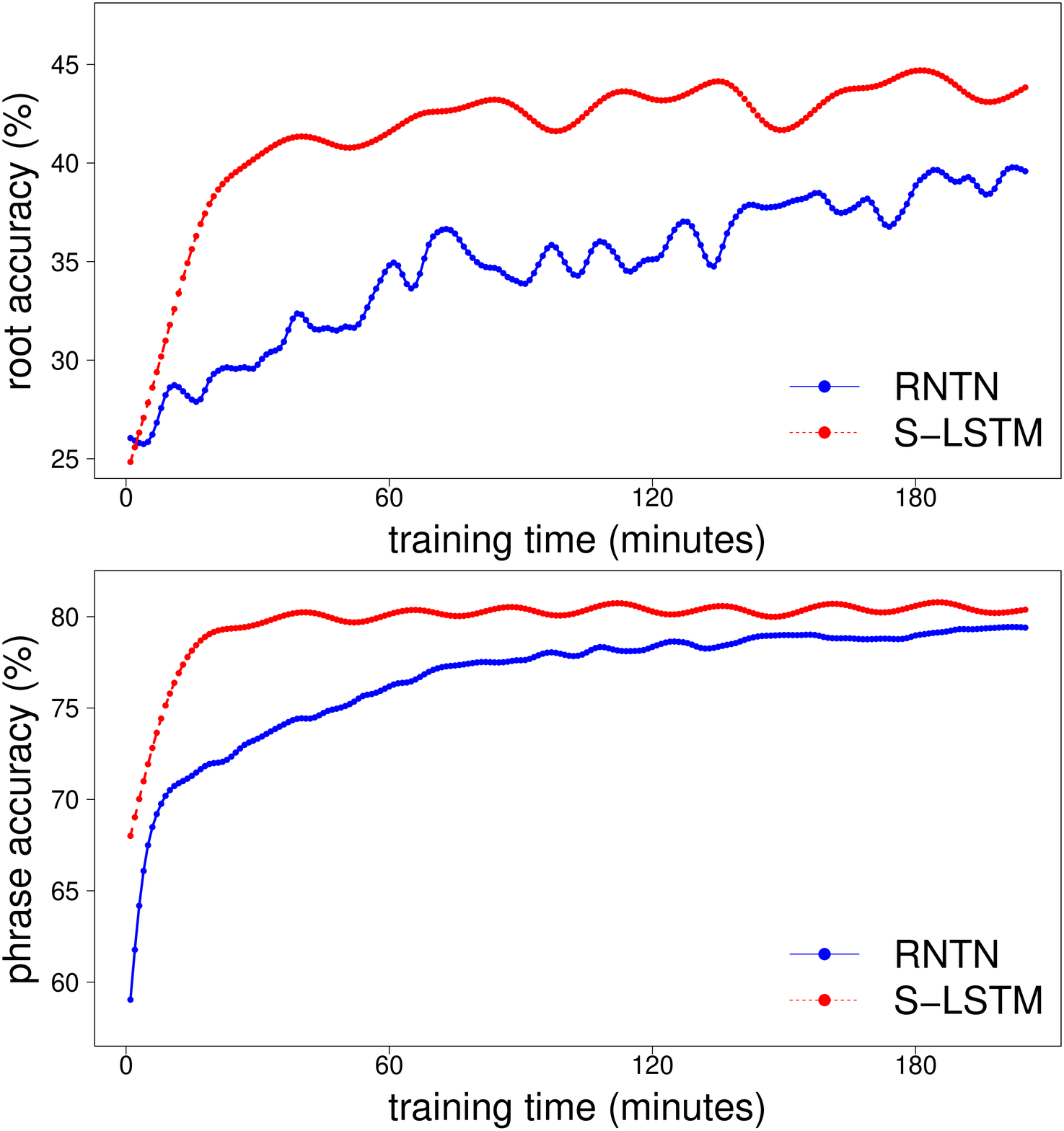}}
\caption{Converging time during training for roots (the upper figure) and for all nodes (the lower figure).}
\label{fig:trainingTime}
\end{figure}

To evaluate the S-SLTM model's convergence behavior, Figure~\ref{fig:trainingTime} depicts the converging time during training. More specifically, we show two sub-figures: one for roots (upper sub-figure) and the other for all phrases (lower sub-figure). From these figures, we can observe that S-LSTM converge faster than the RNTN. For instance, for the phrase-level task, S-LSTM started to converge after about 20 minutes but the RNTN needed over 180 minutes. S-LSTM has much less parameters than RNTN and the forward and backward propagation can be computed efficiently.

\noindent \textbf{More real-life settings}
We further compare S-LSTM with RNTN in two more experimental settings. In the first setting we only keep the training signals at the roots to train S-LSTM and RNTN, depicted as model (1) and (2) in Table \ref{tab:twoSettings}. \textit{ROOT LBLS} besides the model names stands for \textit{root labels}; that is, only the gold labels of the sentence level are used to train the model. In most sentiment analysis circumstances, phrase level annotations are not available: most nodes in a tree are fragments that may not be that interesting; e.g., the fragment \textit{``of a good movie"}~\footnote{Phrase-level sentiment analysis is often defined over a very small subset of phrases of interest, such as in the phrase-level task defined in ~\cite{Wilson05,Mohammad2013}.}. Also, annotating all phrases is expensive. However, these should not be regarded as comments on the value of the Sentiment Tree Bank. Detailed annotations in the tree bank enable much interesting work to be possible, e.g., the study of the effect of negation in changing sentiment~\cite{Zhu2014}.


The second setting, corresponding to model (3) and (4) in Table \ref{tab:twoSettings}, is only slightly different, in which we keep annotation for the tree leafs as well, to simulate that a sentiment lexicon is available and it covers all leafs (words) ($LEAF LBLS$ along the side of the model names stands for \textit{leaf labels}), and so there is no out-of-vocabulary concern. Using real sentiment lexicons is expected to have a performance between the two settings here.

Results in the table show that in both settings, S-LSTM outperforms RNTN by a large margin. When only root labels are used to train the models, S-LSTM obtains an accuracy of 43.5, compared with 29.1 of RNTN. When the leaf labels are also used, S-LSTM achieves an accuracy of 44.1 and RNTN 34.9. All these improvements are statistically significant ($p<0.05$). For the RNTN, without supervising signals from the internal nodes, the composition parameters may not be learned well, potentially because the tensor has much more parameters to learn. On the other hand, through controlling its gates, the S-LSTM shows a very good ability to learn from the trees.   



\begin{table}[h!]
\vskip 0.1in
\caption{Performances of models trained with only root labels (the first two rows) and models that use both root and leaf labels (the last two rows).}
\label{tab:twoSettings}
\vskip 0.1in
\begin{center}
\begin{small}
\begin{sc}
\begin{tabular}{ll}
\noalign{\hrule height 1pt} 
\abovespace\belowspace
Models & roots \\ 
\hline
\abovespace
(1) RNTN (Root Lbls) & 29.1 \\
\belowspace
(2) S-LSTM (Root Lbls) & $43.5$\dag \\ 
\hline
\abovespace
(3) RNTN (Root + Leaf Lbls) & 34.9 \\
\belowspace
(4) S-LSTM (Root + Leaf Lbls) & $44.1$\dag\\ 
\noalign{\hrule height 1pt}
\abovespace
\end{tabular}
\end{sc}
\end{small}
\end{center}
\vskip 0.05in
\end{table}

\begin{figure}[ht]
\vskip 0.2in
\centerline{\includegraphics[width=\columnwidth]{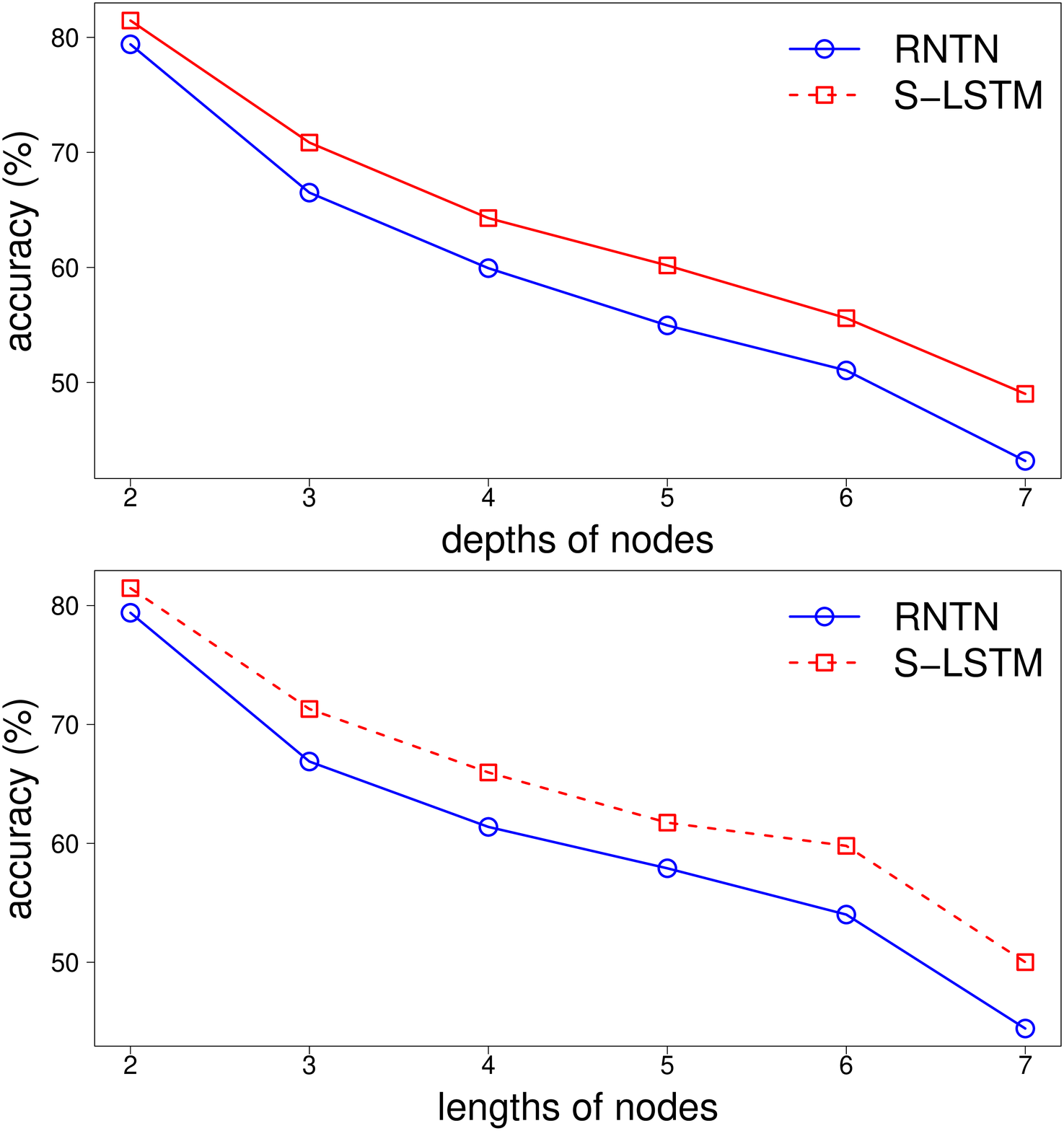}}
\caption{Accuracies at different depths (the upper figure) in the trees, or by different lengths of the phrases (the lower figure).}
\label{fig:errByLevel}
\end{figure}

\noindent \textbf{Performance over different levels of trees}
In Figure \ref{fig:errByLevel}, we further depict the performances of models on different levels of nodes in the trees. In the Figure, the x-axis corresponds to different depths or lengths and y-axis is accuracy. The \textit{depth} here is defined as the longest distance between the root of a phrase and their descendant leafs. The \textit{Length} is simply the number of words of a node, where \textit{depth} is not necessarily to be \textit{length}---e.g., a balanced tree with 4 leafs has different depths than the unbalanced tree with the same number of leafs. The trends of the two figure are similar. In both figures, S-LSTM performs better at all depths, showing its advantages on nodes at depth. As the deeper levels of the tree tend to have more complicated syntax and semantics, S-LSTM can model such more complicated syntax and semantics better.

\noindent \textbf{Explicit structures vs. no structures} 
Some efforts in the literature attempt to learn distributed representation by utilizing input structures when available, and others prefer to assume chain-structured recurrent neural networks can actually capture the structures implicitly though a linear coding process. In this paper, we attempt to give some empirical evidences in our experiment setting by comparing several different models. First, a special case for the S-LSTM model is considered, in which no sentential structures are given. Instead, words are read from left to right and combined in that order. We call it left recursive S-LSTM, or S-LSTM-LR in short. Similarly, we also experimented with a right recursive S-LSTM, S-LSTM-RR, in which words are read from right to left instead. Since for these models, phrase-level training signals are not available---the nodes here do not correspond to that in the original Standford Sentiment Tree Bank, but the roots and leafs annotations are still the same, so we run two versions of our experiments: one uses only training signals from roots and the other includes also leaf annotations. 

\begin{table}[t]
\caption{Performances of models that do not use the given sentence structures. S-LSTM-LR is a degenerated version of S-LSTM that reads input words from left to right, and S-LSTM-RR reads words from right to left.}
\label{tab:flat}
\vskip 0.15in
\begin{center}
\begin{small}
\begin{sc}
\begin{tabular}{ll}
\noalign{\hrule height 1pt} 
\abovespace\belowspace
Models & roots \\ 
\hline
\abovespace
S-LSTM-LR (Root Lbls) & 40.2 \\ 
S-LSTM-RR (Root Lbls) & 40.3 \\ 
\belowspace
S-LSTM (Root Lbls) & 43.5\dag \\ 
\hline
\abovespace
S-LSTM-LR (Root + Leaf Lbls) & 43.1 \\ 
S-LSTM-RR (Root + Leaf Lbls) & 43.2 \\ 
\belowspace
S-LSTM (Root + Leaf Lbls) & 44.1\dag \\ 
\noalign{\hrule height 1pt}
\end{tabular}
\end{sc}
\end{small}
\end{center}
\vskip -0.1in
\end{table}

It can be observed from Table~\ref{tab:flat} that the given parsing structure helps improve the predictive accuracy. In the case of using only root labels, the left recursive S-LSTM and  right recursive S-LSTM have similar performance (40.2 and 40.3, respectively), both inferior to S-LSTM (43.5). When using gold leaf labels, the gaps are smaller, but still, using the parse structure are better. Note that in real applications, where there is no out-of-vocabulary issue (i.e., some leafs are not seen in the sentiment dictionaries), the difference between S-LSTM and the recursive version without using the structures are expected to be between the gaps we observed here.




\section{Conclusions}
We aim to extend the conventional chain-structured long short-term memory to explicitly consider structures.
In this paper we particularly study tree structures, in which the proposed S-LSTM memory cell can reflect the 
history memories of multiple descendants through gated copying of memory vectors. 
The model provides a principled way to 
consider long-distance interplays over the structures. We leveraged the model to learn distributed sentiment representations for texts, and showed that it outperforms a state-of-the-art recursive model by replacing its tensor-enhanced composition layers with the S-LSTM memory blocks.
We showed that the structure information is useful in helping S-LSTM achieve the state-of-the-art performance.

The research community seems to contain two lines of wisdom; one attempts to learn distributed representation 
by utilizing structures when available, and the other prefers to believe recurrent neural networks can actually 
capture the structures implicitly through a linear-chain coding process. In this paper, we also attempt to give some 
empirical evidences toward answering the question. It is at least for the settings of our experiments that
the explicit input structures are helpful in inferring the high-level (e.g., root) semantics.


\bibliography{reference}

\begin{thebibliography}{30}
\providecommand{\natexlab}[1]{#1}
\providecommand{\url}[1]{\texttt{#1}}
\expandafter\ifx\csname urlstyle\endcsname\relax
  \providecommand{\doi}[1]{doi: #1}\else
  \providecommand{\doi}{doi: \begingroup \urlstyle{rm}\Url}\fi

\bibitem[Chater(1992)]{CHATER92}
Chater, N;~Conkey, p.
\newblock finding linguistic structure with recurrent neural networks. lawrence
  erlbaum assoc publ.
\newblock In \emph{in: proceedings of the fourteenth annual conference of the
  cognitive science society. (pp. 402 - 407}, 1992.

\bibitem[Cho et~al.(2014)Cho, van Merrienboer, G{\"{u}}l{\c{c}}ehre, Bougares,
  Schwenk, and Bengio]{Cho:2014}
Cho, Kyunghyun, van Merrienboer, Bart, G{\"{u}}l{\c{c}}ehre, {\c{C}}aglar,
  Bougares, Fethi, Schwenk, Holger, and Bengio, Yoshua.
\newblock Learning phrase representations using {RNN} encoder-decoder for
  statistical machine translation.
\newblock \emph{CoRR}, abs/1406.1078, 2014.

\bibitem[Choi \& Cardie(2008)Choi and Cardie]{Choi08}
Choi, Yejin and Cardie, Claire.
\newblock Learning with compositional semantics as structural inference for
  subsentential sentiment analysis.
\newblock In \emph{Proceedings of the Conference on Empirical Methods in
  Natural Language Processing}, EMNLP '08, pp.\  793--801, Honolulu, Hawaii,
  2008.

\bibitem[Eck \& Schmidhuber(2002{\natexlab{a}})Eck and
  Schmidhuber]{Eck02findingtemporal}
Eck, Douglas and Schmidhuber, J�rgen.
\newblock Finding temporal structure in music: Blues improvisation with lstm
  recurrent networks.
\newblock In \emph{NEURAL NETWORKS FOR SIGNAL PROCESSING XII, PROCEEDINGS OF
  THE 2002 IEEE WORKSHOP}, pp.\  747--756. IEEE, 2002{\natexlab{a}}.

\bibitem[Eck \& Schmidhuber(2002{\natexlab{b}})Eck and
  Schmidhuber]{Eck02learningthe}
Eck, Douglas and Schmidhuber, J�rgen.
\newblock Learning the long-term structure of the blues.
\newblock In \emph{IN PROC. INTL. CONF}, pp.\  284--289, 2002{\natexlab{b}}.

\bibitem[Goller \& Küchler(1996)Goller and
  Küchler]{Goller96learningtask-dependent}
Goller, Christoph and Küchler, Andreas.
\newblock Learning task-dependent distributed representations by
  backpropagation through structure.
\newblock In \emph{In Proc. of the ICNN-96}, pp.\  347--352, Bochum, Germany,
  1996. IEEE.

\bibitem[Graves(2008)]{Graves2008}
Graves, Alex.
\newblock \emph{Supervised sequence labelling with recurrent neural networks}.
\newblock PhD thesis, Technische Universitat Munchen, 2008.

\bibitem[Graves(2012)]{graves2012supervised}
Graves, Alex.
\newblock \emph{Supervised sequence labelling with recurrent neural networks},
  volume 385.
\newblock Springer, 2012.

\bibitem[Graves et~al.(2013)Graves, Mohamed, and Hinton]{Graves:2013}
Graves, Alex, Mohamed, Abdel{-}rahman, and Hinton, Geoffrey~E.
\newblock Speech recognition with deep recurrent neural networks.
\newblock \emph{CoRR}, abs/1303.5778, 2013.

\bibitem[Hammer et~al.(2004)Hammer, Micheli, Sperduti, and
  Strickert]{Hammer04ageneral}
Hammer, Barbara, Micheli, Alessio, Sperduti, Alessandro, and Strickert, Marc.
\newblock A general framework for unsupervised processing of structured data,
  2004.

\bibitem[Hochreiter \& Schmidhuber(1997)Hochreiter and
  Schmidhuber]{Hochreiter1997}
Hochreiter, S. and Schmidhuber, J.
\newblock Long short-term memory.
\newblock \emph{Neural Computation}, \penalty0 (8):\penalty0 1735--1780, 1997.

\bibitem[Irsoy \& Cardie(2014)Irsoy and Cardie]{Irsoy2014}
Irsoy, Ozan and Cardie, Claire.
\newblock Deep recursive neural networks for compositionality in language.
\newblock In Ghahramani, Z., Welling, M., Cortes, C., Lawrence, N.D., and
  Weinberger, K.Q. (eds.), \emph{Advances in Neural Information Processing
  Systems 27}, pp.\  2096--2104. Curran Associates, Inc., 2014.

\bibitem[Kalchbrenner et~al.(2014)Kalchbrenner, Grefenstette, and
  Blunsom]{Kalchbrenner2014}
Kalchbrenner, Nal, Grefenstette, Edward, and Blunsom, Phil.
\newblock A convolutional neural network for modelling sentences.
\newblock \emph{Proceedings of the 52nd Annual Meeting of the Association for
  Computational Linguistics}, June 2014.

\bibitem[Klein \& Manning(2003)Klein and Manning]{Klein2003}
Klein, Dan and Manning, Christopher~D.
\newblock Accurate unlexicalized parsing.
\newblock In \emph{Proceedings of the 41st Annual Meeting on Association for
  Computational Linguistics - Volume 1}, ACL '03, pp.\  423--430, Sapporo,
  Japan, 2003. Association for Computational Linguistics.
\newblock \doi{10.3115/1075096.1075150}.

\bibitem[Liu \& Zhang(2012)Liu and Zhang]{Liu12}
Liu, Bing and Zhang, Lei.
\newblock A survey of opinion mining and sentiment analysis.
\newblock In Aggarwal, Charu~C. and Zhai, ChengXiang (eds.), \emph{Mining Text
  Data}, pp.\  415--463. Springer US, 2012.
\newblock ISBN 978-1-4614-3222-7.
\newblock \doi{10.1007/978-1-4614-3223-4\_13}.

\bibitem[Liwicki et~al.(2007)Liwicki, Graves, Bunke, and
  Schmidhuber]{Liwicki07anovel}
Liwicki, Marcus, Graves, Alex, Bunke, Horst, and Schmidhuber, J�rgen.
\newblock A novel approach to on-line handwriting recognition based on
  bidirectional long short-term memory networks.
\newblock In \emph{In Proceedings of the 9th International Conference on
  Document Analysis and Recognition, ICDAR 2007}, 2007.

\bibitem[Manning \& Sch\"{u}tze(1999)Manning and Sch\"{u}tze]{Manning:1999}
Manning, Christopher~D. and Sch\"{u}tze, Hinrich.
\newblock \emph{Foundations of Statistical Natural Language Processing}.
\newblock MIT Press, Cambridge, MA, USA, 1999.
\newblock ISBN 0-262-13360-1.

\bibitem[Mohammad et~al.(2013)Mohammad, Kiritchenko, and Martin]{Mohammad2013}
Mohammad, Saif~M., Kiritchenko, Svetlana, and Martin, Joel.
\newblock Identifying purpose behind electoral tweets.
\newblock In \emph{Proceedings of the 2nd International Workshop on Issues of
  Sentiment Discovery and Opinion Mining}, WISDOM '13, pp.\  1--9, 2013.

\bibitem[Moilanen \& Pulman(2007)Moilanen and Pulman]{moilanen07}
Moilanen, Karo and Pulman, Stephen.
\newblock Sentiment composition.
\newblock In \emph{Proceedings of RANLP 2007}, Borovets, Bulgaria, 2007.

\bibitem[Pang \& Lee(2005)Pang and Lee]{Pang2005}
Pang, Bo and Lee, Lillian.
\newblock Seeing stars: Exploiting class relationships for sentiment
  categorization with respect to rating scales.
\newblock In \emph{Proceedings of the Annual Meeting of the Association for
  Computational Linguistics}, ACL '05, pp.\  115--124, 2005.

\bibitem[Pang \& Lee(2008)Pang and Lee]{PangL08}
Pang, Bo and Lee, Lillian.
\newblock Opinion mining and sentiment analysis.
\newblock \emph{Foundations and Trends in Information Retrieval}, 2\penalty0
  (1--2):\penalty0 1--135, 2008.

\bibitem[Pinheiro \& Collobert(2014)Pinheiro and Collobert]{pinheiro:2014}
Pinheiro, P. H.~O. and Collobert, R.
\newblock Recurrent convolutional neural networks for scene labeling.
\newblock In \emph{Proceedings of the 31st International Conference on Machine
  Learning (ICML)}, 2014.

\bibitem[Socher et~al.(2011)Socher, Lin, Ng, and Manning]{Socher2011b}
Socher, Richard, Lin, Cliff~C., Ng, Andrew~Y., and Manning, Christopher~D.
\newblock {Parsing Natural Scenes and Natural Language with Recursive Neural
  Networks}.
\newblock In \emph{Proceedings of the 26th International Conference on Machine
  Learning (ICML)}, 2011.

\bibitem[Socher et~al.(2012)Socher, Huval, Manning, and Ng]{Socher2012}
Socher, Richard, Huval, Brody, Manning, Christopher~D., and Ng, Andrew~Y.
\newblock Semantic compositionality through recursive matrix-vector spaces.
\newblock In \emph{Proceedings of the Conference on Empirical Methods in
  Natural Language Processing}, EMNLP '12, Jeju, Korea, 2012. Association for
  Computational Linguistics.

\bibitem[Socher et~al.(2013)Socher, Perelygin, Wu, Chuang, Manning, Ng, and
  Potts]{Socher2013}
Socher, Richard, Perelygin, Alex, Wu, Jean~Y., Chuang, Jason, Manning,
  Christopher~D., Ng, Andrew~Y., and Potts, Christopher.
\newblock Recursive deep models for semantic compositionality over a sentiment
  treebank.
\newblock In \emph{Proceedings of the Conference on Empirical Methods in
  Natural Language Processing}, EMNLP '13, Seattle, USA, 2013. Association for
  Computational Linguistics.

\bibitem[Starzyk et~al.()Starzyk, Member, and
  He]{Starzyk_anticipation-basedtemporal}
Starzyk, Janusz~A., Member, Senior, and He, Haibo.
\newblock Anticipation-based temporal sequences learning in hierarchical
  structure.

\bibitem[Sutskever et~al.(2014)Sutskever, Vinyals, and Le]{Sutskever:2014}
Sutskever, Ilya, Vinyals, Oriol, and Le, Quoc~V.
\newblock Sequence to sequence learning with neural networks.
\newblock \emph{CoRR}, abs/1409.3215, 2014.

\bibitem[Vinyals et~al.(2014)Vinyals, Toshev, Bengio, and Erhan]{Vinyals:2014}
Vinyals, Oriol, Toshev, Alexander, Bengio, Samy, and Erhan, Dumitru.
\newblock Show and tell: {A} neural image caption generator.
\newblock \emph{CoRR}, abs/1411.4555, 2014.

\bibitem[Wilson et~al.(2005)Wilson, Wiebe, and Hoffmann]{Wilson05}
Wilson, Theresa, Wiebe, Janyce, and Hoffmann, Paul.
\newblock Recognizing contextual polarity in phrase-level sentiment analysis.
\newblock In \emph{Proceedings of the Conference on Human Language Technology
  and Empirical Methods in Natural Language Processing}, HLT '05, pp.\
  347--354, Stroudsburg, PA, USA, 2005. Association for Computational
  Linguistics.
\newblock \doi{10.3115/1220575.1220619}.

\bibitem[Zhu et~al.(2014)Zhu, Guo, Mohammad, and Kiritchenko]{Zhu2014}
Zhu, Xiaodan, Guo, Hongyu, Mohammad, Saif, and Kiritchenko, Svetlana.
\newblock An empirical study on the effect of negation words on sentiment.
\newblock In \emph{Proceedings of ACL}, Baltimore, Maryland, USA, June 2014.

\end{thebibliography}
\bibliographystyle{icml2015}

\end{document}